\documentclass{article}

\usepackage{arxiv}
\usepackage[utf8]{inputenc} 
\usepackage[T1]{fontenc}    
\usepackage{hyperref}       
\usepackage{url}            
\usepackage{booktabs}       
\usepackage{amsfonts}       
\usepackage{nicefrac}       
\usepackage{microtype}      
\usepackage{graphicx}
\usepackage[square,authoryear]{natbib}
\usepackage{doi}
\usepackage{xcolor}
\usepackage{xspace}

\title{An Automated Length-Aware Quality Metric for Summarization}

\author{ 
	\href{}{
		\hspace{1mm}
		Andrew D.~Foland}
	\thanks{This work is not affiliated with Leidos, Inc.
	}\\ 
    Sonnetiq\\
	\texttt{andrew.foland@sonnetiq.com} \\
}

\date{}

\renewcommand{\shorttitle}{\textit{NOIR} Metric}

\hypersetup{
pdftitle={arxiv submission},
pdfsubject={CS.ai},
pdfauthor={Andrew D. ~Foland},
pdfkeywords={Summarization, Metrics, Semantic Measures, Sentence Embeddings},
}

\begin{document}
\maketitle

\begin{abstract}
This paper proposes NOrmed Index of Retention (NOIR), a quantitative objective metric for evaluating summarization quality of arbitrary texts that relies on both the retention of semantic meaning and the summary length compression. 
This gives a measure of how well the recall-compression tradeoff is managed,
the most important skill in summarization.
Experiments demonstrate that NOIR effectively captures the token-length / semantic retention tradeoff of a summarizer and correlates to human perception of sumarization quality. 
Using a language model-embedding to measure semantic similarity, it provides an automated alternative for assessing summarization quality without relying on time-consuming human-generated reference summaries. 
The proposed metric can be applied to various summarization tasks, offering an automated tool for evaluating and improving summarization algorithms, summarization prompts, and synthetically-generated summaries.
\end{abstract}


\section{Introduction}
Quantitative methods for summarization capability evaluation today still often have drawbacks for the era of high-rate automated task completion via LLM's.
Several methods have been devised for quantitative measures of summarization quality, as recently reviewed in \citep{2023arXiv230504853R} and \citep{fabbri_summeval_2021}. 

\begin{figure}
	\centering
	\includegraphics[height=6cm]{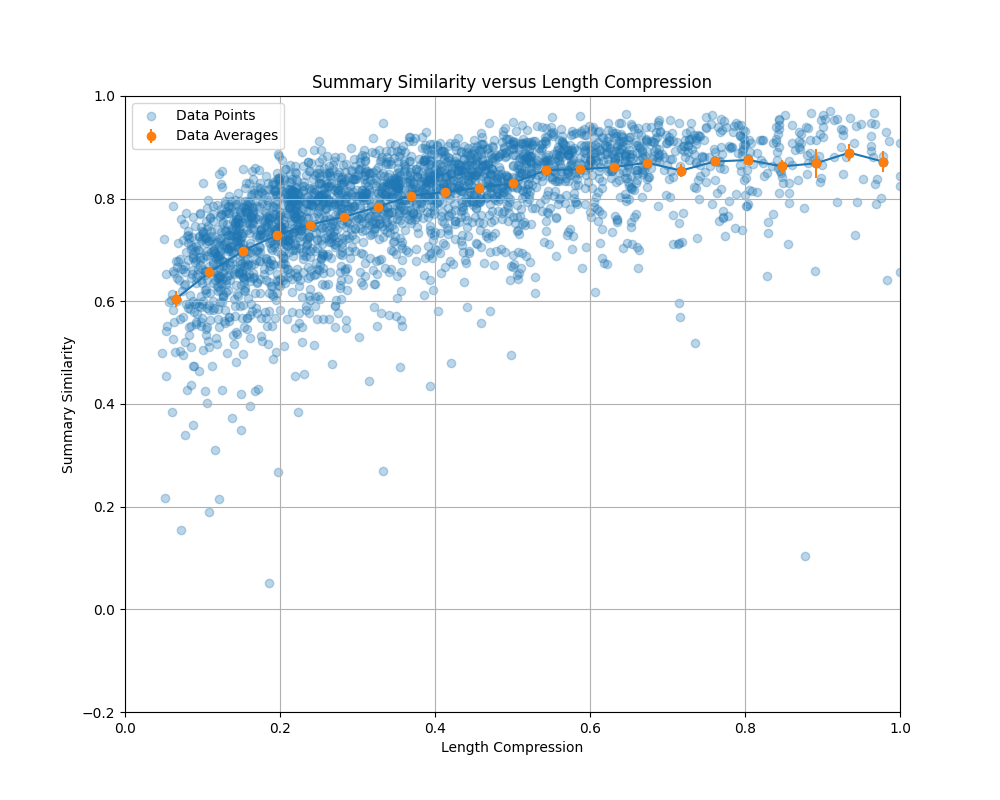}
	\caption{Summary embedding similarity to parent document similarity as a function of length compression.  As the summarizer uses fewer tokens, semantic similarity is reduced.  The quality of a summarizer can be measured by how quickly it falls off as a function of token compression.}
	\label{fig:similarityvslength}
\end{figure}

Skill at summarization is generally demonstrated by maximizing content retention for a given level of summary compression\footnote{Famously, the short story commonly attributed to Hemingway: "For sale: baby shoes, never worn."}.  
Indeed, as seen in figure \ref{fig:similarityvslength}, it becomes (unsurprisingly) increasingly difficult to retain all semantic content as the length of the summary is reduced.
This figure provides the motivation and observation for this paper: that better summarization techniques should fall higher, and worse ones lower, on this plot of semantic similarity versus compression; and that one scalar metric should be capable of expressing where any technique falls on this plot.

Surprisingly, standard methods do not generally compare candidate summaries to the original text to be summarized.
Most methods focus on precision or recall compared to a reference dataset of summaries \textit{rather than to the text to be summarized itself}; yet these do not necessarily align to human skill in summarization; for instance recall score can easily be increased by writing a longer summary--contrary to the spirit of summarization.  

We therefore wish to incorporate an explicit comparison to the parent text, including a measure of token length compression, into our evaluation of summarization quality.
Quantifying a good balance is desirable for automated summary generation, for instance by large language models (LLM's).

We aim to capture this skill as a measurable property of a summarization process, based only on the parent text and the candidate summary.
To our knowledge NOIR is the first method to explicitly take into account the compression / recall tradeoff in a text/candidate context evaluating summarization quality.
It is also, to our knowledge, the first method to measure that tradeoff motivated in a form with no free parameters or \textit{ad hoc} terms.

We emphasize that the precise measure of semantic similarity, subject to numerical assumptions in form described below, can be changed without necessarily changing the nature of the NOIR metric.
As semantic measures improve over time, NOIR is straightforwardly generalized to use them in place of the measures proposed in the present communication.
We demonstrate this in the course of the present report by using an alternate embedding model.

\section{Contributions}

In this paper we contribute:

\begin{enumerate}
	\item An objective, automated, reference-free, no-free-parameters summarization metric comparing parent text and summary directly that rewards conciseness explcitly and is usable on arbitrary texts and summaries
	\item A dataset of LLM-generated candidate summaries of 456 different texts, where the summaries vary in length and quality
	\item A demonstration that the metric captures well the tradeoff of a summarizer between token length and semantic retention
	\item Evidence for only weak embedder-dependence of the metric under particular numerical conditions as to the form of the similarity measure
	\item Evidence for correlation of metric output values to human perception of summarization quality	
\end{enumerate}

\section{Background}

\subsection{Previous Work}
As discussed in \citep{2023arXiv230504853R}:

\begin{quote}
Most commonly, summarization systems are evaluated on automated metrics. ROUGE \citep{lin_rouge_2004} in particular has a long-standing history in the field and measures the lexical overlap between reference summaries and generated summaries. More recent metrics such as BertScore \citep{zhang_bertscore_2019} and BARTScore \citep{yuan_bartscore_2021}, which are better at capturing semantic equivalence, are also becoming increasingly established. 
\end{quote}

One such is the ROUGE metric \citep{lin_rouge_2004}, which compares the generated summary to a reference summary and calculates the overlap between the two.
The process of comparing a summary to reference summaries (and not the parent paragraph) using the ROUGE metric is described as "standardized" for summarization evaluation in \citep{fabbri_summeval_2021}.
Another attempt is PRIMERA \citep{xiao_primera_2022}, which evaluates the content coverage and structural similarity of the generated summary to the reference summary.  
Many other methods are described in detail in \citep{fabbri_summeval_2021}.
\\
A method comparing the text and candidate summary in a similar spirit to this proposal is SUPERT \citep{supert}; however SUPERT does not take into account length, and relies on automated generation of extractive reference summaries from the text for comparison to the summary.  Likewise, the BLANC \citep{vasilyev-etal-2020-fill} technique dispenses with the reference documents without taking length into account.

Another method recently published is \citep{guo-vosoughi-2023-length} which attempts to correct for a length bias in summarization measures; this implicitly achieves some of the aims of the present NOIR metric.  It is differently motivated, however, and does not explicitly reward skilled compression in the way that the NOIR metric does.

\section{The Methodology}

We wish to devise a metric for summarization quality. 
It is to be expected that there is always a tradeoff between faithfulness to the original text's full meaning, and the length of the summary.
A good summary is one that retains as much fidelity as possible given its length.
We formulate a metric that is based on a ratio of semantic retention to token compression.

\subsection{Overview}
Consider an idealized summarizer that compresses the token count $T$ by a multiplicative compression factor $k$ at each step and reduces its semantic similarity by a multiplicative degradation factor $D$ at each step.  
High summarization quality corresponds to $k$ small and $D$ large, so the quantity $\frac{D}{k}$ (or any similar function monotonically increasing in $D$ and monotonically decreasing in $k$) forms a measure of the summarization quality of this idealized summarizer. After $N$ compressions, we have $T_N$ tokens left:

\begin{equation}
	\label{eq:tokencount}
    T_N = T_0 k^N,
\end{equation}

and semantic degradation $D$ is given by

\begin{equation}
	\label{eq:degradationsteps}
	D_N = D_0 D^N = D^N
\end{equation}

for $D_0=1$, which we are free to scale.

We would like a metric on the observables $D^N$ and $\frac{T_N}{T_0}$ that is monotonically increasing in $D$ and decreasing in $k$, when $k$ and $N$ are unknown. 
The metric $M_{NOIR}$ that does this is:

\begin{equation}
M_{\rm NOIR} = \frac{\ln\frac{T_N}{T_0} }{\ln D_N } =  \left(  \frac{\ln k}{\ln D} \right)
\end{equation}

Note that is indeed increasing in $D$ and decreasing in $k$ as desired, for $D$, $k$ < 1.
We keep it in the form $\frac{\ln k}{\ln D}$ as it is more suggestive than the reduction to $\log_D k$.

Note particularly that by taking this form, we do not need to know what $N$ or $k$ is--we measure a quality metric value directly rather than model parameters of a postulated summarization process.  
The postulated idealized summarizer model serves only to motivate and illuminate the form, which stands or falls on its own usefulness and correlation to perceived quality.

The form also means that the quality as measured is continuous--we are not restricted to integer $N$.
Finally, we note that the metric goes smoothly to negative values--in the case where the summary is longer than the original text.  
Even in the perfect-semantic preservation case, one would expect this to have negative utility, as captured by the metric.

\subsection{Measurement of Semantic Degradation}

While counting tokens $T_N$ is very straightforward, how are we to measure the semantic degradation $D$?

We propose to measure it by evaluating the similarity that
derives from a semantic sentence embedder \citep{hill-etal-2016-learning}, \citet{DBLP:journals/corr/abs-2002-10957}, \citet{DBLP:journals/corr/SutskeverVL14}, \citep{muennighoff2022mteb}.

A sentence embedding is a dense vector representation of a sentence, which captures its semantic meaning in a continuous vector space. It is a form of word embedding, which has been extended to handle entire sentences. The concept of sentence embeddings has evolved from the need to move beyond word-level representations, as words can have different meanings based on context. The history of sentence embeddings can be traced back to the early 2000s, with foundational work on word embeddings like Word2Vec \citep{mikolov2013efficient} and GloVe \citep{pennington-etal-2014-glove}. The development of recursive neural networks and, later, transformer-based models allowed for the creation of more sophisticated sentence embeddings, capable of capturing long-range dependencies and nuanced meanings. Notable techniques include contextualized embeddings from models like BERT \citep{DBLP:journals/corr/abs-1810-04805} and RoBERTa \citep{DBLP:journals/corr/abs-1907-11692}. These advancements have significantly improved the performance of natural language processing tasks such as semantic textual similarity, paraphrase detection, and machine translation.

It is important to choose a sentence embedder and similarity metric with the following features: 

\begin{itemize}
	\item Semantically identical meanings should have a similarity measure of 1
	\item Semantically unrelated meanings should have a similarity measure of 0
\end{itemize}

A cosine-similarity measure of a sentence embedding vector will often have these features: 

\begin{equation}
	D \equiv \hat{v}_{summ} \cdot \hat{v}_{text}.
\end{equation}

Note that should better measures than embeddings or cosine become available in the future, they can be fully incorporated into NOIR subject to the numerical conditions of limiting at 0 and 1, and functioning approximately multiplicatively.

While these embedders are often referred to as "sentence embeddings", they can take as input multiple sentences.  Token counts up to 512 are common, and even up to 4096 in some cases not unheard of.  This allows paragraphs to be semantically embedded in the space.

\subsection{Motivation for Cosine Similarity as Degradation Measure}

Why should we expect degradation in cosine similarity to be multiplicative?

Degradation by $D$ in cosine space corresponds to moving by some angle (i.e., $\frac{\pi}{2}-\arccos(D)$). 
In the embedding space, this is approximately moving by some "random" direction and length in embedding space by that angle.

If (by Ansatz) the lengths are constant at each step, but the angles are fixed (with random direction), then the net random step is a length in some direction, along some distance $\ell_{\perp}$ and $\ell_{\parallel}$.
Crucially, in high dimensions the random $\ell_{\perp}$ directions at different steps will be orthogonal to one another \citep{randomwalk}. 
So the accumulated orthogonal component grows like a random walk\footnote{Note that the $\sqrt{N}$ behavior continues to hold in high dimensions.} of $N$ orthogonal steps--aka like $\sqrt{N}$.  

\begin{figure}
	\centering
    \includegraphics[height=6cm]{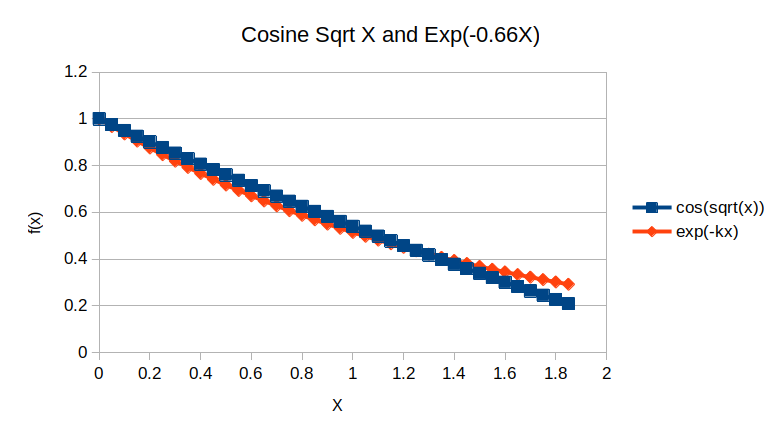}
	\caption{Demonstration of similarity of $\cos \sqrt{x} $ and $e^{-0.66x}$ over the range from $\cos \sqrt{x}=1$ to $\cos \sqrt{x} = 0.2$.  The coefficient value of 0.66 is found by a least-squares fit to the cosine over the range.  The functional closeness motivates the choice of cosine similarity as a multiplicative semantic degradation measure.}
	\label{fig:expcos}
\end{figure}

For small angles, the angle itself grows linearly with the orthogonal component.
Hence the relevant function looks like $\cos{\alpha \sqrt{N}}$, where $\alpha$ is controlled by the amount of parallel and perpendicular components per step.
This function looks strikingly like an exponential over most of its range--hence the Ansatz of multiplicative loss--as shown in figure \ref{fig:expcos}.
The range from $\cos \sqrt{x}=1$ to $\cos \sqrt{x} = 0.2$ is the relevant one for summaries\footnote{A true degradation operator would not take on negative values, suggesting that there may be a better candidate function of the embedding vectors than the cosine similarity with which to evaluate the semantic similarity.  However as we will see it serves with the properties we desire over this range.}, as random text pairings can give rise to semantic similarities as large as 0.2.

To the extent that one can extend the discrete "summarization step" to a continuous "summarization operator" which can be applied at arbitrary strength, with 0-strength corresponding to the identity operator, the exponential function (multiplicative Ansatz) correponds to a linear and real operator generator.

\subsection{Sentence Embeddings}


Many sentence embedders use only a subset of the feature space \citep{badembeddings}, resulting in particularly the second condition (0 similarity for unrelated texts) being violated.
We choose an embedder for which we can show both conditions hold.

We show that two commonly-used sentence embedders, both of which satisfy both conditions, yield broadly similar results when evaluating the summarization metric over a common dataset.


\subsection{Summarization}

Summarization is a common NLP task. 
Given the recent interest in large language models (LLM's) and automated text generation, interest is renewed in automated summzarization.
Given this, it is cumbersome for human-dependent reference texts to form the backbone of summarization evaluation.

In this paper we target an objective metric of summarization based on semantic similarity.
This metric does not depend on human judgment or creation of a good reference text.

\subsection{Content and Length Variability}

It would be unsurprising if summarization quality were content-dependent.  
For the purposes of this paper, we work to include a broad array of content and topics.

As sentences are quantized at the word level, it would not be unexpected that summarization quality measured by the metric would be dependent on the text length, particularly for low values of the text length.  We investigate the dependence of the metric on this length.

\subsubsection{Length Correlations}

If the sentence embedding model encodes (either explicitly, or for instance as a distributed principal component) the input sequence length, then apparent semantic differences in summarization could in principle be a combination of perfect semantic agreement coupled with an encoded length difference.  
Similar questions have been investigated previously, e.g., in \citep{finegrained}.
We investigate a paraphrase-based test and other vector-component level tests to investigate this possibility.

\section{Experimental Setup}

We describe how we evaluate the utility of the proposed NOIR metric.  
As described previsouly, the metric stands or falls on its own usefulness and correlation to perceived quality.
We therefore wish to demonstrate the following:

\begin{itemize}
	\item Independence of embeddings from input length
	\item That the similarity metric we use is near 1 for semantically similar texts and near 0 for semantically unrelated texts
    \item That the summarization quality metric shows only weak dependence on the sentence embedding model (once the key 0-to-1 requirement is met)
	\item That the summarization quality metric is near 0 when the "summarization" of a text is unrelated to the text
	\item That the summarization quality metric metric forms a compact distribution away from 0 for summaries of different lengths known to derive from a parent unsummarized text
	\item That the summarization quality metric operates as expected not only on summaries, but also hierarchically on summaries of summaries
	\item That the metric correlates to known quantitative measures and qualitative perceptions of summarization quality.
\end{itemize}

\subsection{Datasets}

\subsubsection{Summarization Dataset}
For investigations, we use the MultiRC2 (R2) dataset \citep{multirc2}.
The dataset consists of 456 paragraphs drawn from many different topics, styles, and sources.
These are described in \ref{tab:summarizationdataset}, as taken from \citep{multirc2}.
We convert each paragraph into a single clean text prior to embedding.
Typical paragraph lengths are 200-500 words.

\begin{table}
	\caption{Topics and sources in the summarization dataset, taken from \cite{multirc2}.}
	\centering
	\begin{tabular}{lll}
		\toprule
		\multicolumn{2}{c}{Sources}                   \\
		\cmidrule(r){1-2}
		Category     & Sources     & Approx. $\%$ \\
		\midrule
		News  	 	& CNN, NYT, WSJ & 15$\%$    \\
		Articles	& Wikipedia		& 10$\%$    \\
		Articles on law and justice	& Ide and Suderman 2006 & 10$\%$\\
		Articles on history and anthropology 	& Ide et al 2009	& 5$\%$ \\
		Elementary School Science Textbooks	& www.ck12.org	& 20$\%$\\
		9/11 Reports	& Ide and Suderman, 2006	& 10$\%$ \\
		Fiction		& Project Gutenberg, Children's Stories, CMU Movie Summaries & 30$\%$\\
		
		\bottomrule
	\end{tabular}
	\label{tab:summarizationdataset}
\end{table}

\subsubsection{Paraphrase Dataset}

For paraphrases, we desire a dataset with paragraphs that can usefully be paraphrased and are of typical paragraph size. 
Ideally they would be paragraphs no LLM has seen before.
For this purpose we use \citep{randomparagraphs} to seed a few-shot prompt for generation of random paragraphs.
We obtain 6 seed paragraphs spanning different settings, styles, and points of view.  
We seed these as few-shot "random paragraphs" and the LLM  (\texttt{LoneStriker\_dolphin-2.5-mixtral-8x7b-6.0bpw-h6-exl2-2}) generates new random paragraphs.
We harvest 490 paragraphs of typically 40-100 words randomly generated in this way. 

We then proceed to paraphrase these 490 parapgraphs.  
We seed the 6 seed paragraphs, along with 8 manual paraphrases each, as a few-shot paraphrase exercise and the LLM  (\texttt{LoneStriker\_dolphin-2.5-mixtral-8x7b-6.0bpw-h6-exl2-2}) generates new paraphrase paragraphs as a continuation ({\textit{nb}, not as an instruction-following chat format.)

The fewshot continuation prompt is structured as an instructional text with examples, wherein the examples are of concise and verbose paraphrases of exemplar texts.  
The instructional heading of the prompt is

\begin{itemize}
	\item \textbf{Paraphrase Fewshot prompt header}: "\# What is Paraphrasing?
	Paraphrasing refers to the process of rephrasing a piece of text or information while retaining its original meaning. It is an essential tool 
	used by writers, editors, and students to convey the same message in different ways, making it easier for readers to understand and remember the information. Paraphrasing is important because it helps to improve writing skills, avoid plagiarism, and enhance comprehension by presenting information in a more concise and clear manner. Additionally, paraphrasing can be used to tailor information to different audiences, making 
	it more accessible and engaging for readers.  Paraphrasing can be verbose and wordy, expanding the length of the passage, or concise and succinct, shortening the passage." \textit{Followed by 6 example texts, each with 8 example paraphrases of varying lengths.}
\end{itemize}

\subsection{Sentence Embedding}

For sentence embedding, we use the \texttt{all-MiniLM-L6-v2} \citep{DBLP:journals/corr/abs-2002-10957} embedding model.
This model is chosen for its widespread use and good properties.  
It can embed up to 512 tokens, typically about 400 words.
The most important good property that is has, shown in figure \ref{fig:randomsimilarity}, is that random pairs of text drawn from our text dataset show cosine similarity centered near 0.  
Many embeddings show much larger values of cosine similarity (often centered in the range 0.4-0.7) for random pairings of text samples.
This may indicate \citep{badembeddings} that such models are using only a subspace of the full embedding dimension.

\begin{figure}
	\centering
	\includegraphics[height=6cm]{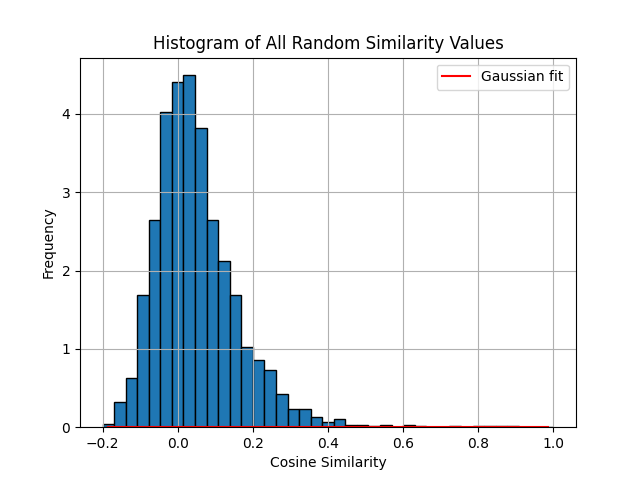}
	\caption{Distribution of semantic similarity for random pairings of paragraphs with the \texttt{all-MiniLM-L6-v2} sentence embedding.}
	\label{fig:randomsimilarity}
\end{figure}

\subsection{Language Model and Prompts}

We use a language model to generate summaries of our text dataset.
The language model we use is \texttt{LoneStriker\_Mixtral-8x7B-Instruct-v0.1-5.0bpw-h6-exl2}, a quantized version of a high-performing open-weights Mixture-of-Experts model.
For each text, the model was hierarchically invoked and prompted to produce a summary, first of the original text, then of the summary of the original text, then of the summary of the summary.  
The model was prompted as follows:

\begin{itemize}
	\item \textbf{Original text}: "Provide a concise 200-word summary of the key information in the following text.  Provide the summary in the same voice and tense as the original.  Do not add anything else.  Provide only the requested 200-word summary:"
	\item \textbf{First summary}: "Provide a short, concise summary, of no more than 75 words, of the following text.  Provide the summary in the same voice, tense, and view as the input text.  Reduce the length significantly.  Do not add anything else.  Provide only the requested 75-word short summary:"
	\item \textbf{Second summary}: "Provide a one-sentence summary of the following text, retaining only the most important information.  Reduce the length significantly.  Provide the summary in the same voice and tense as the original text.  Do not add anything else.  Provide only a short, concise, one-sentence summary:"
\end{itemize}

\subsection{Note on Token Count}

When we give token counts in this paper, we give them based on the \texttt{tiktoken} token counter package from OpenAI \citet{tiktoken}.
This is neither the tokenizer used by our embedders (which are custom to each embedder) nor our LLM's (which use the llama tokenizer \citet{touvron_llama_2023}).  However, the various tokenizers tend to give similar counts within the precision of this paper ($\pm 20\%$) and are not expected to impact the conclusions, as token counts in the paper are used mainly to form compression ratios in which overall token count scale factors cancel.

\subsection{Test Summarizations}

To investigate summarization, we summarize the text hierarchically as described previously.  
Each summary is embedded with the sentence embedder.  
For each summary, the ratio of output summary tokens to input text tokens $\frac{T_{summ}}{T_{text}}$ is calculated for the denominator of the $M_{NOIR}$ metric.  
The cosine similarity $\hat{v}_{text}\cdot\hat{v}_{summ}$ is evaluated as $D$ for the numerator of the $M_{NOIR}$ metric.

We show in figure \ref{fig:summary_similarities} that the similarity of a text with its summary is typically near 1.

\begin{figure}
	\centering
	\includegraphics[height=6cm]{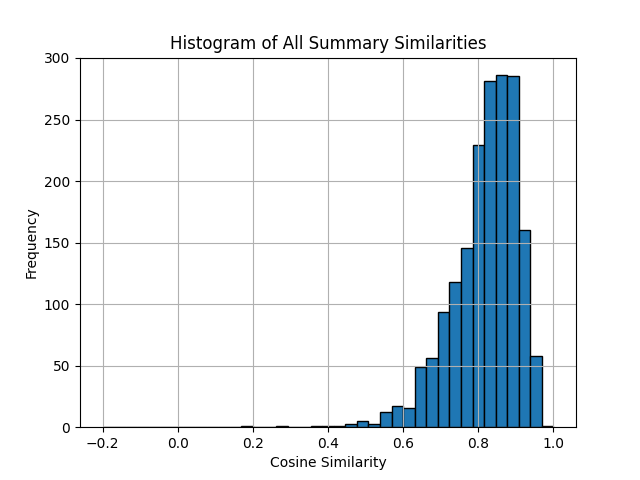}
	\caption{Distribution of cosine similarity for pairings of paragraphs with their summaries.  The distribution is peaked near 1.0.}
	\label{fig:summary_similarities}
\end{figure}

\subsection{Null Summarizations}

To demonstrate that the cosine similarity metric is near zero for unrelated texts, we randomly pair (exluding the correct pairings) texts with the summaries of other texts.  
The resulting distribution is shown in figure \ref{fig:randomsimilarity}, peaking near 0.

\subsection{Paraphrase Tests}

We investigate the length dependence and high-similarity behavior of the sentence embedding using a dataset of paraphrases.  
The paraphrases are generated by the \texttt{LoneStriker\_dolphin-2.5-mixtral-8x7b-6.0bpw-h6-exl2-2} language model.  
The model was prompted few-shot with six paragraph examples of generating (for each paragraph) four concise and verbose paraphrases.

Four "concise" and four "verbose" paraphrases are generated for each input. 
Due to imperfections in the LLM, even the "concise" samples are often longer than the original text.
We review all paraphrases manually, and all are acceptable examples of paraphrasing of the original.

\subsubsection{Semantic Similarity Distribution}

In figure \ref{fig:summary_similarities}
we showed that paraphrase similarity to input text is very high, typically larger than 0.9.  
In the high (368) dimensional space of the embedding, cosine similarity of 0.9 indicates very strong alignment.  

In figure \ref{fig:paraphrase_normlength}, 
we show a scatter plot of semantic similarity as a function of normalized paraphrase length.  Over a range near unchanged paraphrase length, similarity is high and flat, falling off at more extreme differences.
It is important to know whether the falloff is due truly to the small semantic differences that can exist between paraphrases (and which surely exist more when conditioned on large length difference) or whether there might be elements of the embedding that correlate directly to length of the embedded text. 
Such elements would confound measurement of an image quality metric combining semantic information with length information, over-incorporating the length information.

\begin{figure}
	\centering
	\includegraphics[height=6cm]{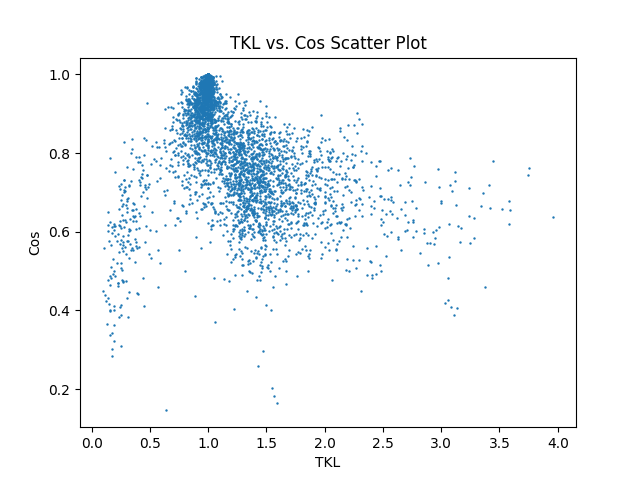}
	\caption{Distribution of cosine similarity for paragraphs and their paraphrases, as a function of normalized paraphrase length $\frac{\ell_{para}}{\ell_{text}}$. }
	\label{fig:paraphrase_normlength}
\end{figure}

\subsubsection{Evaluation of Text Length Correlations}

We evaluate whether the embedding directly encodes the input length.
Over all 4410 examples, we examine correlation of each of the 368 vector dimensions to both length and the normalized length (length divided by length of the original.) 
We believe normalized length is a better measure for this test to isolate the variations of semantically similar texts.
The observed correlations are shown in figure \ref{fig:corrnormlength}.

For normalized length, we find the distribution of correlations is centered at zero with a standard deviation of the distribution of approximately 0.2.  
We find we cannot reject the null hypothesis that no component is correlated with length.  
We find no components with absolute value of correlation greater than 0.45.
We conclude that single components with weak correlation are unlikely to induce significant shifts in the semantic cosine similarity due to token length variation alone.

Similarly, we see no pattern of strong correlations either with the raw length, or with any of the vector components of a principal components decomposition.

\begin{figure}
	\centering
	\includegraphics[height=6cm]{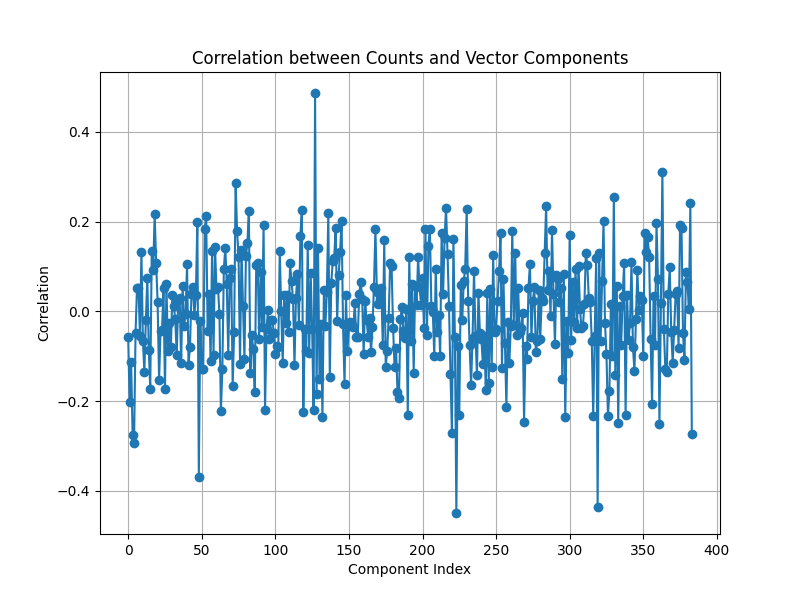}
	\caption{Observed correlations between 368 individial embedding vector components and the normalized length of the input.  No correlation patterns strong enough to materially alter cosine similarity due to length alone are evident.}
	\label{fig:corrnormlength}
\end{figure}

\subsection{Code and Dataset Availability}
All of the code and datasets used for this paper can be found at \texttt{https://github.com/afoland/NOIR}

\section{Results and Analysis}

\subsection{Performance Evaluation}

First we show that summarization quality is very low and tightly clustered near 0 when the summary is not a summary of the parent paragraph, in Figure \ref{fig:mNOIR_random}.

\begin{figure}
	\centering
	\includegraphics[height=6cm]{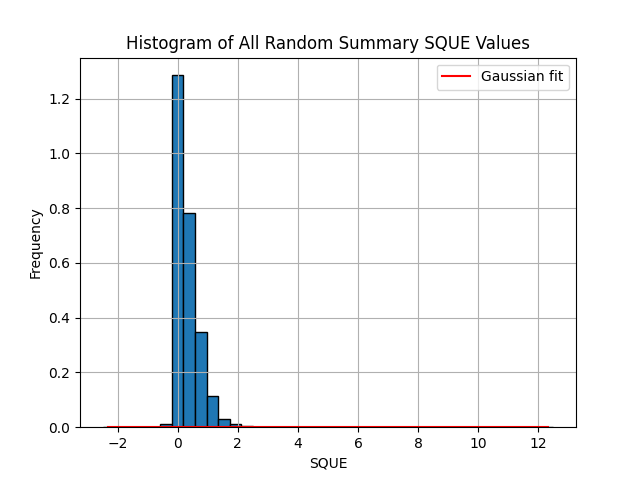}
	\caption{Distribution of $M_{NOIR}$ quality for random pairings of paragraphs with summaries of other paragraphs.  The mean value is 0.35, close to 0.}
	\label{fig:mNOIR_random}
\end{figure}

Next we show that the summarization quality is distinctively better and clustered significantly away from zero for true summarizations.
The distribution of $M_{NOIR}$ is shown in figure \ref{fig:mNOIR}.
The mean of the distribution 4.55$\pm$0.06 is separated by over two standard deviations of the distribution (and by many\footnote{While we report the fit uncertainty value of 0.06 on the mean, we believe this may understate the uncertainty somewhat, as there may be correlations between, for instance, the $M_{NOIR}$ evaluation of a text and summary, and between the summary and its own summary, due to being in similar parts of semantic embedding space. It is beyond the scope (and likely beyond the useful point of value) of the present report to precisely determine these correlations and a more accurate value of the fit uncertainty.} standard errors on the mean).

The meaning of $M_{NOIR}=4.55$ is that for each halving of a token length in summarization, the semantic degradation is $\ln{2}/4.55$, or a multiplicative factor of 0.86.  
Note the good agreement of this value with the average cosine similarity (i.e., degradation) at 0.5, 0.25 ($0.86^2=0.74$), and 0.125 ($0.86^3=0.63$) in figure \ref{fig:similarityvslength}.
This implies the metric captures well the tradeoff curve of the summarizer in length versus semantic retention.

\begin{figure}
	\centering
	\includegraphics[height=6cm]{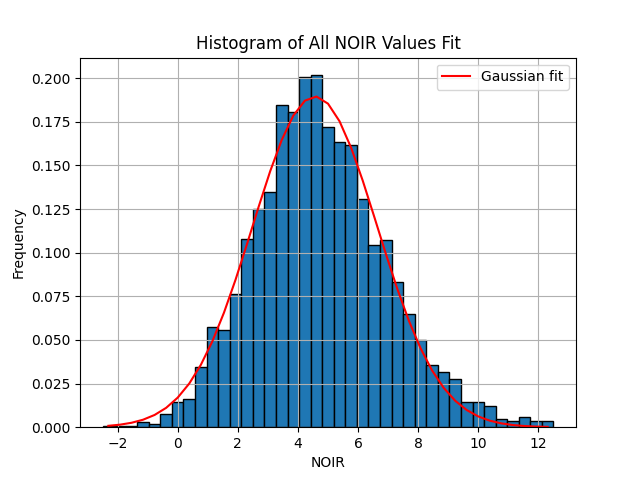}
	\caption{Distribution of $M_{NOIR}$ quality for summaries, summaries of summaries, and summaries of summaries of summaries.  Also shown is a gaussian fit to the distribution.  The fitted mean value is 4.55$\pm$0.06, with a Gaussian $\sigma$ of 2.08.  It is well-separated from the values for random pairings of paragraphs and other summaries.}
	\label{fig:mNOIR}
\end{figure}

Finally, in figure \ref{fig:mNOIR_trend} we show there is no meaningful length-dependent trend in quality.  We find no evidence for trend on average, with an upper limit of no more than 10\% variation per 500 tokens.

\begin{figure}
	\centering
	\includegraphics[height=6cm]{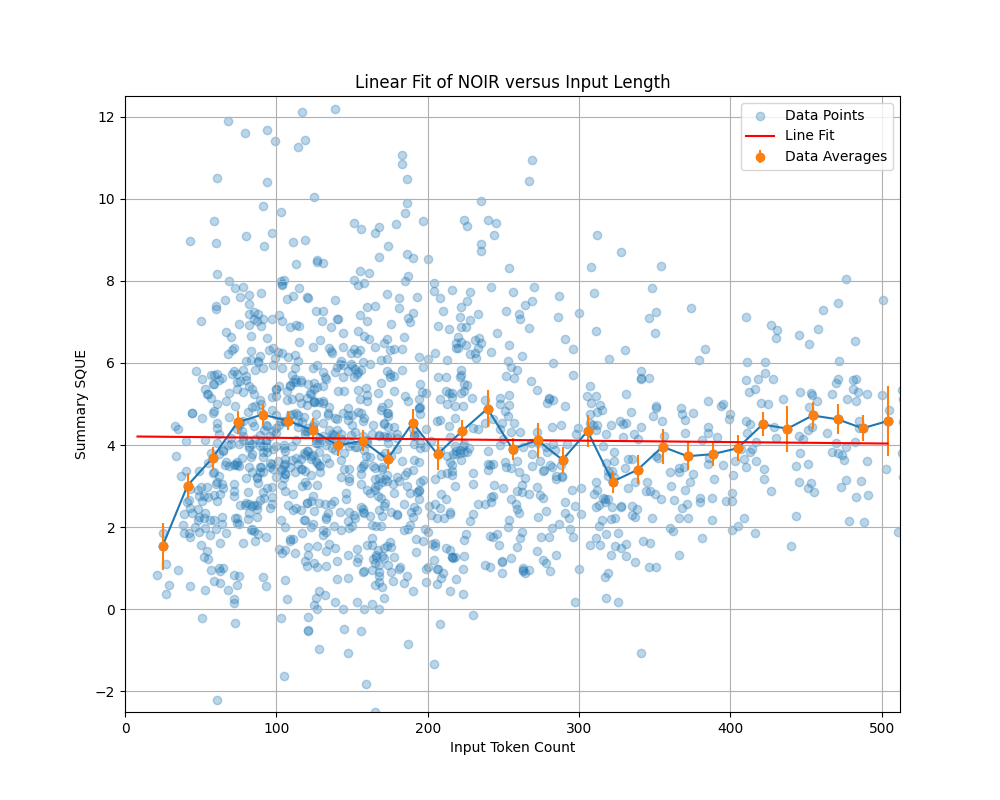}
	\caption{Distribution of $M_{NOIR}$ against input text length (in tokens). In addition to the scatterplot, the average $y$ value ($M_{NOIR}$ value) is shown, together with its standard error on the mean.  Also shown is a linear fit to the trend.  The fitted slope is 0.0003$\pm$0.0005, or at most about 0.40 units of $M_{NOIR}$ quality (about 10\% variation) per 500 tokens}
	\label{fig:mNOIR_trend}
\end{figure}

\subsection{Hyperparameter Studies}

\subsubsection{Form of the NOIR Expression}

As motivated, the expression for NOIR has no free parameters.  
However, as a test, we evaluate the case where the NOIR expression is modified by raising the numerator to a power:

\begin{equation}
	M_{\rm NOIR} = \frac{-\ln|\frac{T_N}{T_0}|^p}{\ln D}
\end{equation}

where $p$ is a power in the range $\left[0,2\right]$, and the absolute values are necessary for fractional $p$ as the compression factor is nearly always negative.

In the limit $p \rightarrow 0$, we recover ${\rm NOIR} \rightarrow \hat{v}_{summ} \cdot \hat{v}_{text}$ (i.e., pure semantic simlarity), with no dependence on the length compression.
As $p \rightarrow 2$, the NOIR value becomes increasingly tolerant of semantic loss for highly-compressed summaries, leading to increasing overlap with random pairings of summaries.

We evaluate the separation of the random NOIR distribution and the summary NOIR distribution, $\frac{\mu_{summ}-\mu_{random}}{\sqrt{\sigma_{summ}^2+\sigma_{random}^2}}$.
	The resulting distribution, shown in figure \ref{fig:powerseparation]}, has a broad maximum, peaking at 1.0.  We conclude that 1.0, as motivated, is empirically an acceptable point of operation.

\begin{figure}
	\centering
	\includegraphics[height=6cm]{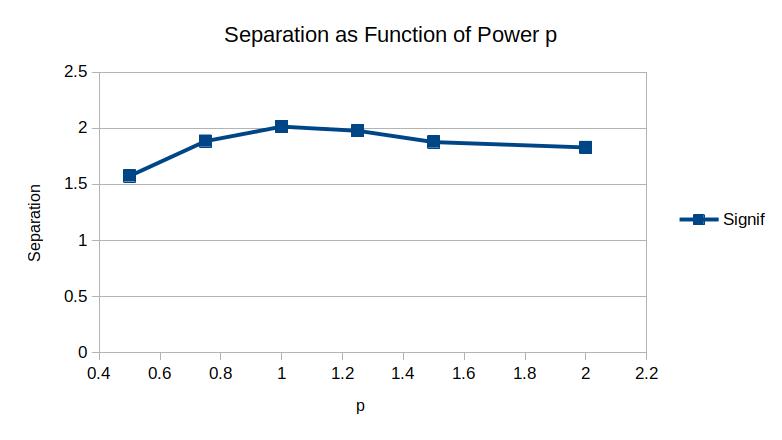}
	\caption{$M_{NOIR}$ distribution separation (in standard deviations of the distribution) between random pairings of text, and true summaries, as a function of the power $p$ applied to the denominator of NOIR.  The separation peaks at 1.0, the originally-motivated value of $p$, and we conclude this is an empirically acceptable value.}
	\label{fig:powerseparation]}
\end{figure}

\subsubsection{Using a Different Embedder}

We repeat the analysis using the commonly 
used \texttt{all-mpnet-base-v2}  
\citep{mpnet}.
The identical texts and summaries are used.

As shown in \ref{fig:mpnet}, this embedder satisfies the numerical requirements for similar and dissimilar texts.
It also shows a similar-shaped falloff of semantic retention as a function of token compression.
This embedder finds modestly higher NOIR values, that are generally correlated ($\rho=0.55$) with the NOIR values found using \texttt{all-MiniLM-L6-V2}.
The value of NOIR found is 5.8$\pm$0.09 with a standard deviation of 2.9;
the distribution's separation from random texts (NOIR=0) is nearly identical (i.e. 2$\sigma$) to that of the default embedding model.

The higher NOIR value implies that more of the semantic content important to \texttt{all-mpnet-base-v2} is being retained than the semantic content important to \texttt{all-MiniLLM-L6-v2}.  
This is an example of the known differences among embedders in their sensitivity to different semantic aspects.

\begin{figure}
	\centering
	\includegraphics[height=3cm]{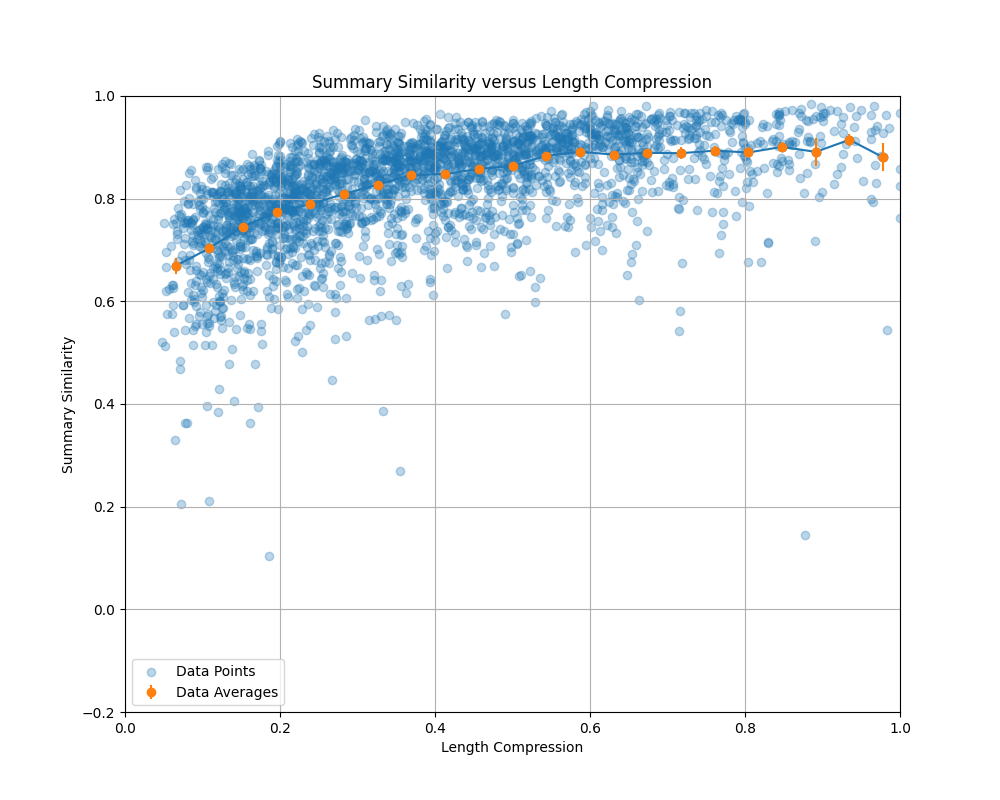}
	\includegraphics[height=3cm]{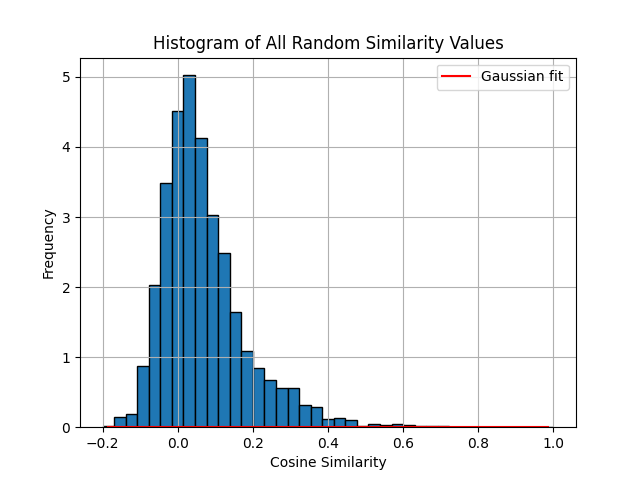}
	\includegraphics[height=3cm]{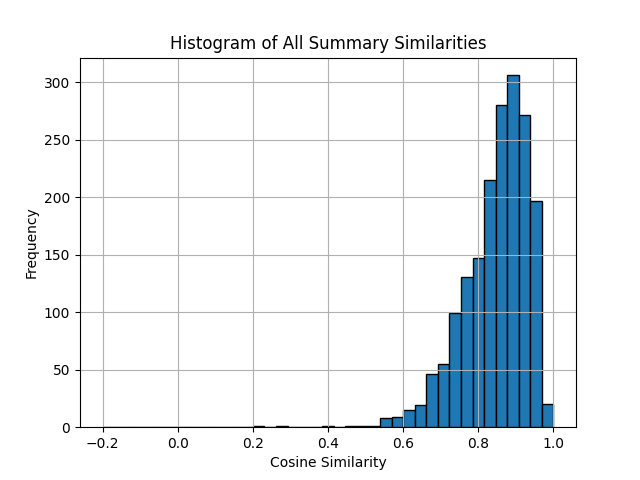 }
	\includegraphics[height=3cm]{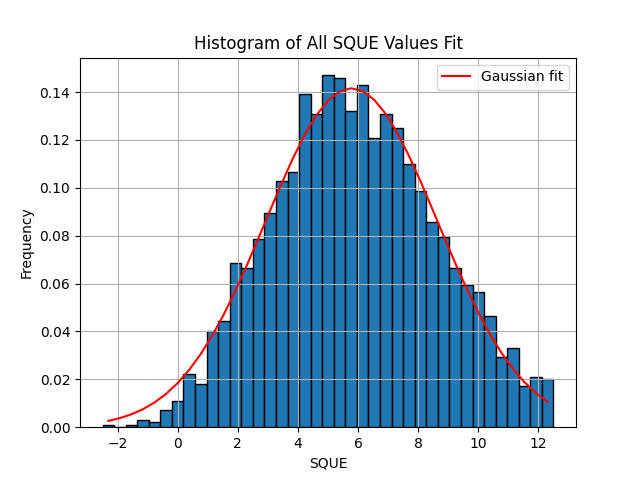}
	\caption{Analysis of the summary dataset using another embedder, \texttt{all-mpnet-base-v2}.  From left to write: similarity as a function of token cmopression; similarity for random unrelated text pairings; similarity for summary pairs; distribution of observed $M_{NOIR}$ values.  The NOIR values are modestly higher for this embedder.}
	\label{fig:mpnet}
\end{figure}

\subsection{Comparison to Human Evaluation}

As described previously, human evaluation of summary quality considers both semantic retention and compression.  
We wish to evaluate NOIR's correlation to human evaluation.  
We select 3 texts in each of 5 bins of compression([0.2,0.3],[0.3,0.4],[0.4,0.5],[0.5,0.6],[0.6,0.8]).  
The 3 texts selected  are the texts nearest to the 10\%, 50\%, and 90\% percentile
in NOIR for that compression bin.
The fifteen texts are presented blindly and in random order to the human evaluator for rank ordering in summarization quality.

The resulting rank correlations are shown in figure  .
Note that the top two summaries by NOIR were correctly ranked in order as the top
two summaries by the human evaluator.  
The correlation coefficient of the rank ordering is 0.36.
This is significantly larger than has been found for the widely-used ROUGE metric \citep{5071230} (Table 1) and larger than most
standard evaluation methods \citep{supert} (Table 1).
We conclude that NOIR with this embedding correlates with human perception of summarization quality.

\begin{figure}
	\centering
	\includegraphics[height=6cm]{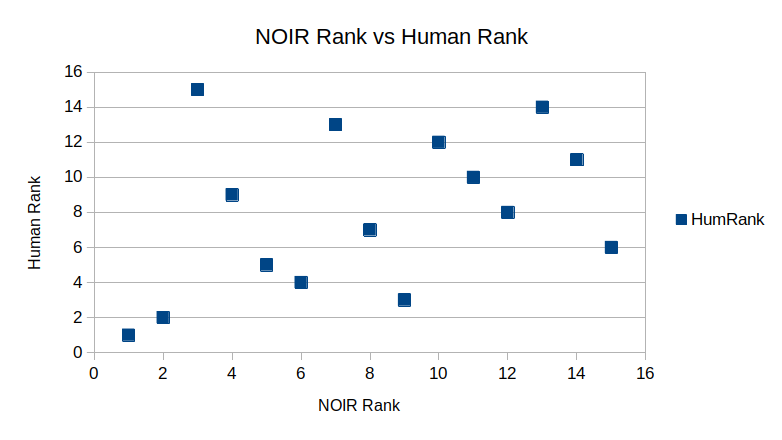}
	\caption{Human rank-ordering of test-summary pair summarization quality as a function of NOIR metric rank for the pair.  The correlation is +0.36, demonstrating that NOIR correlates to human perception of summarization quality.}
	\label{fig:humaneval}
\end{figure}

\subsection{Efficiency and Objectivity}

The entire summarization dataset, consisting of 1824 embeddings of approximately 360,000 tokens, is evaluated in less than 60 seconds on a Intel Xeon 28-core CPU.
We conclude that evaluation of the metric is very fast, significantly faster than most LLM summarization processes.

The evaluation is objective insofar as it is fully automated and does not depend on any subjective evaluation or human input.
The only inputs are the (arbitrary) text and its candidate summary.

\section{Conclusion}

\subsection{Summary of Contributions}

In this paper we have demonstrated 

\begin{enumerate}
\item An automated, objective, reference-free, parameter-free summarization metric that directly compares parent text and summary, that explicitly rewards conciseness and is usable on arbitrary texts and summaries
\item That the metric captures well the tradeoff of a summarizer between token length and semantic retention
\item That the metric definitively distinguishes between summaries of texts, and random summary pairings of shorter lengths but unrelated meanings (the latter having quality near zero)
\item Indications of modest embedder-dependence of the metric (conditioned on numerical preconditions being met )
\item Correlation of metric output values to human perception of summarization quality
\end{enumerate}

Harnessing sentence embedding techniques and a length normalization, NOIR moves beyond traditional lexical overlap methods and subjective reference summary requirements. 

By comparing the semantic embeddings of both input documents and generated summaries, NOIR offers an efficient, objective, and content-coverage-focused evaluation method. 

\subsection{Potential Impact and Future Work}

In the world of LLM-automated content generation and task completion, an automated, high-speed summarization quality metric is likely to play an important role in development.  

Future work might include investigation of topic dependence and vulnerability/stability of the metric under a wider array of embeddings, particularly at long input lengths.

\subsection{Real-World Applications}
The proposed NOIR metric can be utilized to enable high-quality synthetic summary generation by filtering automatically generated summaries. Summarization algorithms can be employed to create a pool of candidate summaries for a given input document.  The NOIR metric can be applied to each automatically generated summary, calculating the semantic embedding closeness scores. By setting a predefined threshold based on experimental analysis or domain knowledge, the system can filter out summaries with lower-quality scores, retaining only those that surpass the threshold.

The filtered high-quality summaries can then be utilized for various purposes, such as feeding back into the training process of summarization models, offering users a selection of high-quality summaries to choose from, or being used as reference summaries for further evaluation.

Periodic updates of the threshold and re-evaluation of generated summaries can help refine the summary generation process, leading to continuous improvement in summarization algorithm performance and generated summary quality. This filtering mechanism based on the NOIR metric ensures objective, efficient, content coverage-focused evaluation without the need for time-consuming human-generated reference summaries.

The metric can also be used when crafting prompts for summarization, as feedback to indicate what prompts are likely to be most successful in generating good summaries for a given task.

\bibliographystyle{unsrtnat}
\bibliography{references}

\end{document}